\documentclass[11pt]{article}
\usepackage[final]{acl}

\usepackage{times}
\usepackage{latexsym}

\usepackage[T1]{fontenc}

\usepackage[utf8]{inputenc}

\usepackage{microtype}

\usepackage{inconsolata}

\usepackage{graphicx}

\usepackage{booktabs, multirow, adjustbox, makecell}
\usepackage{amsmath}
\usepackage{xspace}
\usepackage{listings}
\definecolor{codebg}{RGB}{246,248,250}
\definecolor{codeframe}{RGB}{208,215,222}
\title{A Composable Evaluation System for\\ Reproducible Omni-Modal Foundation Model Evaluation}

\author{
  \textbf{Hodong Lee\textsuperscript{1,2}},
  \textbf{Sanghee Park\textsuperscript{1,3}},
  \textbf{Dohoon Ryu\textsuperscript{1}},\\
  \textbf{Jungwhan Kim\textsuperscript{1}},
  \textbf{Junyeob Kim\textsuperscript{4}\thanks{Work done while at NAVER Cloud AI.}},
  \textbf{Soyoon Kim\textsuperscript{1}},
  \textbf{Geewook Kim\textsuperscript{1,3}\thanks{Corresponding author.}}
\\
\\
  \textsuperscript{1}NAVER Cloud AI,
  \textsuperscript{2}Korea University,
  \textsuperscript{3}KAIST AI,
  \textsuperscript{4}Seoul National University
\\
  \small{\texttt{\{hodong.lee, sang.hee.park, dh.ryu, jungwhan.kim, soyoon.kim, gw.kim\}@navercorp.com}}
\\
  \small{\texttt{juny116@europa.snu.ac.kr}}
}

\begin{document}
\maketitle

\begin{abstract}
Building an omni-modal foundation model means evaluating it across text, image, video, and audio.
Excellent evaluation toolkits exist for each modality, but their inference engines, prompt conventions, and metric implementations are mutually incompatible, so practitioners end up maintaining separate environments for every toolchain and still struggle to compare results across them.
\textbf{OmniEvaluator} grew out of this need in our own model development: rather than reimplementing benchmarks, it connects existing inference engines and curated evaluation libraries at a higher level, exposing four inference backends, four evaluation frameworks, and over a thousand benchmarks through a single interface.
Every run is recorded as an artifact capturing the full configuration for exact reproduction, and results flow into a shared dashboard for cross-model comparison.
A federated mode shares GPU inference servers across concurrent evaluations, and a built-in \emph{verifier}, small enough to run on CPU, gives a score that varies far less across engines and prompts than rule-based scoring under configuration mismatch, matching cost-efficient commercial LLM judges without their recurring API cost.
The system, demo video, and dashboard are publicly available.\footnote{\url{https://github.com/naver-ai/omni-evaluator}}
\end{abstract}

\begin{figure}[t!]
  \centering
  \includegraphics[width=0.92\columnwidth]{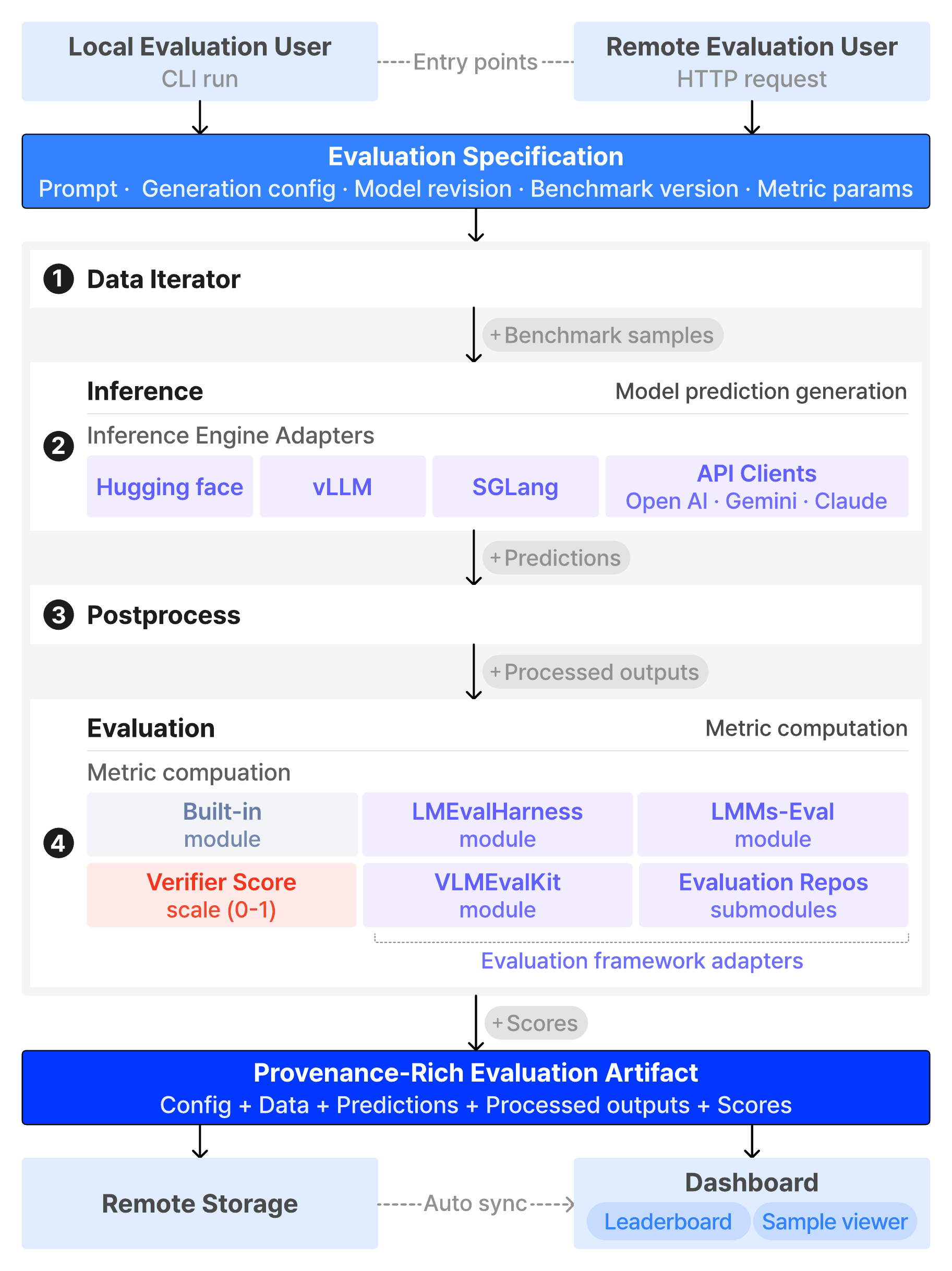}
  \caption{
  \textbf{OmniEvaluator architecture.}
  Inference and evaluation engines are composed through a unified intermediate schema, enabling modular combination of any supported engine pair.
  Users can run evaluations locally via CLI or submit requests to a remote evaluation server, both producing provenance-rich evaluation artifacts.
  }
  \label{fig:architecture}
\end{figure}

\section{Introduction}

Foundation models are increasingly \emph{omni-modal}: a single model accepts text, image, video, and audio as input~\cite{geminiteam2025geminifamilyhighlycapable,openai2024gpt4ocard,xu2025qwen25omnitechnicalreport,fu2025vita,navercloudhyperclovaxteam2026hyperclovax8bomni}.
Developing such a model requires evaluating it on benchmarks from every modality it supports.
Evaluation infrastructure, however, has not kept pace.
Vision--language models are usually evaluated with VLM-specific toolkits~\cite{10.1145/3664647.3685520}, text models with language-model harnesses~\cite{lintang_sutawika_2026_20122284}, and audio or video capabilities with scripts written for individual benchmarks.
Each toolchain has its own prompt conventions, preprocessing steps, and scoring code.

This fragmentation causes two recurring problems for research and production teams alike.
First, no single environment covers an omni-modal model.
Each toolkit pins its own dependencies; per-toolkit virtual environments avoid version conflicts but still leave every toolkit with its own interface, configuration format, and output layout, all reconciled by hand.
In practice, modalities that are hard to set up often simply go unmeasured.
Second, scores reported for ``the same benchmark'' often disagree.
Different libraries use different prompt templates, decoding parameters, and evaluator versions~\cite{biderman2026lessonstrenchesreproducibleevaluation,alzahrani-etal-2024-benchmarks}, and even metrics with the same name, such as WER~\citep{park-etal-2024-automatic} or ANLS~\citep{peer2025anlsuniversaldocument}, can hide different normalization rules.
As a result, scores are hard to compare across papers, and sometimes even across runs within a team (we quantify this in Table~\ref{tab:engine_comparison}).
As a further consequence, results become scattered across machines, and GPU resources are split across users rather than pooled.

Our goal is not to build yet another isolated benchmark suite, but to tie together what the community already builds and maintains, so that a single model can be evaluated across every modality it supports, under one configuration, with every score reproducible.
We present \textbf{OmniEvaluator}, a composable evaluation system that combines existing inference engines and curated evaluation libraries as building blocks (Figure~\ref{fig:architecture}).
Internally, a shared intermediate schema standardizes the four stages of evaluation: data iteration, inference, postprocessing, and metric computation.
Connecting $N$ inference engines to $M$ evaluation frameworks directly would require $N \times M$ pairwise integrations.
With the schema in between, each engine and each framework needs only one thin adapter, $N + M$ in total, and any engine can then be paired with any framework.
Every run also produces a \emph{provenance-rich artifact} recording the full configuration (prompt template, generation parameters, model revision, benchmark version, and metric settings), so a score can be reproduced rather than merely reported.

\begin{table}[t]
\centering
\caption{
\textbf{Supported inference and evaluation engines.}
Each engine is integrated via a single adapter to a unified schema.
\textbf{\# Bench.} counts the benchmarks accessible per evaluation
framework at the time of writing; the built-in engine covers audio and other benchmarks
unavailable elsewhere.}
\label{tab:supported_engines}
\resizebox{\columnwidth}{!}{%
\begin{tabular}{@{}llc@{}}
\toprule
\textbf{Category} & \textbf{Supported Engine} & \textbf{\# Bench.} \\
\midrule
\multirow{4}{*}{Inference}
  & HuggingFace Transformers~\cite{wolf-etal-2020-transformers} & -- \\
  & vLLM~\cite{10.1145/3600006.3613165} & -- \\
  & SGLang~\cite{NEURIPS2024_724be447} & -- \\
  & Off-the-shelf API clients (OpenAI, Google, Anthropic) & -- \\
\midrule
\multirow{4}{*}{Evaluation}
  & built-in & 182 \\
  & \texttt{lm-eval-harness}~\cite{lintang_sutawika_2026_20122284} & 1986 \\
  & \texttt{lmms-eval}~\cite{zhang-etal-2025-lmms} & 416 \\
  & \texttt{VLMEvalKit}~\cite{10.1145/3664647.3685520} & 375 \\
\bottomrule
\end{tabular}
}%
\end{table}

On top of this pipeline, OmniEvaluator provides:
\begin{itemize}
  \setlength{\itemsep}{2pt}
  \item \textbf{Unified omni-modal evaluation}: one installation and one interface covering four inference backends, four evaluation frameworks, and over a thousand benchmarks across text, image, video, and audio (Table~\ref{tab:supported_engines})---spanning a substantial portion of the benchmarks reported in the omni-modal tech reports published to date (Table~\ref{tab:tech_report_coverage})---with any backend able to run any framework's benchmarks. A federated mode further shares inference servers across concurrent evaluations for efficient GPU use (\S\ref{sec:federated_evaluation}).

  \item \textbf{Integrated dashboard}: an interactive view that gathers per-modality results in one place, shows coverage gaps at a glance, and supports cross-model and cross-checkpoint comparison for model selection (Figure~\ref{fig:dashboard}).

  \item \textbf{Built-in verifier}: a small model, light enough to run on CPU, that judges whether a prediction is semantically correct.
  Its normalized score varies far less across engines and prompts than rule-based scoring under configuration mismatch, reducing the need for costly API judges (Table~\ref{tab:engine_comparison}, \S\ref{sec:verifier}).

\end{itemize}

OmniEvaluator was used in the development of HyperCLOVA X 8B Omni~\cite{navercloudhyperclovaxteam2026hyperclovax8bomni} and is publicly available with a live demo, demo video, and evaluation dashboard.


\begin{table}[t]
\centering
\caption{\textbf{Benchmark coverage of OmniEvaluator.}
Number of benchmarks reported in each representative omni-modal foundation model's tech report that OmniEvaluator supports.}
\label{tab:tech_report_coverage}
\resizebox{\columnwidth}{!}{%
\begin{tabular}{@{}lc@{}}
\toprule
\textbf{Technical Report} & \textbf{Covered / Total} \\
\midrule
HyperCLOVAX-SEED-Omni-8B~\citep{navercloudhyperclovaxteam2026hyperclovax8bomni} & 28 / 28  \\
MiniCPM-o 4.5~\citep{cui2026minicpmo45realtimefullduplex}        & 49 / 62 \\
Qwen3.5-Omni~\citep{qwenteam2026qwen35omnitechnicalreport} & 54 / 72 \\
Qwen2.5-Omni~\citep{xu2025qwen25omnitechnicalreport} & 47 / 56 \\
Qwen3-Omni~\citep{xu2025qwen3omnitechnicalreport} & 50 / 65 \\
\bottomrule
\end{tabular}
}%
\end{table}

\section{Related Work}
\label{sec:related}

Evaluation tools have matured within each modality; what is missing is a way to evaluate one model across all of them under a single, consistent setup.

\paragraph{Text evaluation.}
HELM~\cite{liang2023holistic} popularized holistic evaluation; \texttt{lm-eval-harness}~\cite{biderman2026lessonstrenchesreproducibleevaluation} provides reusable task implementations and documents common pitfalls; BIG-bench~\cite{srivastava2023beyond} and the Open LLM Leaderboard~\cite{open-llm-leaderboard-v2} broadened coverage and showed centralized evaluation at scale.
Evalverse~\cite{kim-etal-2024-evalverse} is closest to our design, unifying several text evaluation frameworks behind one interface, but at the time of writing it does not extend beyond text.

\begin{table}[t]
  \centering
  \caption{\textbf{Native vs.\ verifier score across evaluation engines and prompt conditions.} Each cell: \texttt{lmms-eval}/\texttt{VLMEvalKit} on the same outputs, under benchmark-specific (\emph{Sp.}) and uniform (\emph{Un.}) prompts; $[0,100]$ scale, $\Delta{=}$max$-$min. \emph{Sp.}\ uses each framework's own format-constraining instruction (e.g.\ ``Answer the question using a single word or phrase.''), while \emph{Un.}\ replaces it with a single format-agnostic instruction for all benchmarks; full prompts in Table~\ref{tab:prompt_conditions}.}
  \label{tab:engine_comparison}
  \renewcommand{\arraystretch}{1.0}\scriptsize
  \setlength{\tabcolsep}{2pt}
\begin{tabular}{@{}lll cccc c@{}}
    \toprule
    & & & \multicolumn{2}{c}{\textbf{lmms-eval}} & \multicolumn{2}{c}{\textbf{VLMEvalKit}} & \\
    \cmidrule(lr){4-5}\cmidrule(lr){6-7}
    \textbf{Bench.} & \textbf{Model} & \textbf{Score} & \textbf{Sp.} & \textbf{Un.} & \textbf{Sp.} & \textbf{Un.} & \textbf{$\Delta$} \\
    \midrule
    \multirow{4}{*}{GQA}
      & \multirow{2}{*}{Qwen2.5-Omni-3B} & Native   & 61.6 & 28.1 & 70.4 & 29.3 & 42.3 \\
      &                                  & Verifier & 64.8 & 66.4 & 64.7 & 66.6 & \phantom{0}1.9 \\
      \cmidrule(l){2-8}
      & \multirow{2}{*}{Qwen2.5-Omni-7B} & Native   & 60.9 & 28.5 & 70.1 & \phantom{0}0.9 & 69.2 \\
      &                                  & Verifier & 63.3 & 63.9 & 63.1 & 66.0 & \phantom{0}2.9 \\
    \midrule
    \multirow{4}{*}{MMStar}
      & \multirow{2}{*}{Qwen2.5-Omni-3B} & Native   & 57.1 & 51.0 & 54.2 & 53.2 & \phantom{0}6.1 \\
      &                                  & Verifier & 55.2 & 53.7 & 54.4 & 53.5 & \phantom{0}1.7 \\
      \cmidrule(l){2-8}
      & \multirow{2}{*}{Qwen2.5-Omni-7B} & Native   & 62.7 & 47.9 & 62.1 & 61.1 & 14.8 \\
      &                                  & Verifier & 61.5 & 48.0 & 62.3 & 61.5 & 14.3 \\
    \midrule
    \multirow{4}{*}{OCRBench}
      & \multirow{2}{*}{Qwen2.5-Omni-3B} & Native   & 82.2 & 82.1 & 25.6 & 25.6 & 56.6 \\
      &                                  & Verifier & 84.3 & 84.3 & 84.6 & 84.2 & \phantom{0}0.4 \\
      \cmidrule(l){2-8}
      & \multirow{2}{*}{Qwen2.5-Omni-7B} & Native   & 80.5 & 83.2 & 25.2 & 25.2 & 58.0 \\
      &                                  & Verifier & 86.0 & 84.9 & 85.1 & 84.8 & \phantom{0}1.2 \\
    \midrule
    \multirow{4}{*}{POPE}
      & \multirow{2}{*}{Qwen2.5-Omni-3B} & Native   & 88.8 & 59.2 & 87.6 & 87.9 & 29.6 \\
      &                                  & Verifier & 88.8 & 88.8 & 91.3 & 91.1 & \phantom{0}2.5 \\
      \cmidrule(l){2-8}
      & \multirow{2}{*}{Qwen2.5-Omni-7B} & Native   & 88.4 & \phantom{0}0.0 & 87.6 & 86.4 & 88.4 \\
      &                                  & Verifier & 88.4 & 87.1 & 91.5 & 87.8 & \phantom{0}4.4 \\
    \midrule
    \multirow{4}{*}{RealWorldQA}
      & \multirow{2}{*}{Qwen2.5-Omni-3B} & Native   & 64.6 & 64.4 & 62.6 & 63.8 & \phantom{0}2.0 \\
      &                                  & Verifier & 64.7 & 64.6 & 62.6 & 63.8 & \phantom{0}2.1 \\
      \cmidrule(l){2-8}
      & \multirow{2}{*}{Qwen2.5-Omni-7B} & Native   & 69.2 & 69.9 & 69.7 & 69.5 & \phantom{0}0.7 \\
      &                                  & Verifier & 69.3 & 70.1 & 69.7 & 69.5 & \phantom{0}0.8 \\
    \bottomrule
  \end{tabular}
\end{table}

\paragraph{Vision evaluation.}
VLMEvalKit~\cite{10.1145/3664647.3685520} unifies evaluation across a wide range of vision--language models, and VHELM~\cite{NEURIPS2024_fe2fc7dc} extends holistic evaluation to the vision--language setting.
Benchmarks such as MMMU~\cite{Yue_2024_CVPR}, MMBench~\cite{10.1007/978-3-031-72658-3_13}, and Video-MME~\cite{Fu_2025_CVPR} standardize \emph{what} to measure, but \emph{how} they are run (prompts, answer parsing, scoring) still differs from framework to framework.

\paragraph{Multi-modal and audio evaluation.}
LMMs-Eval~\cite{zhang-etal-2025-lmms} extends coverage beyond image--text.
For audio, dedicated toolkits such as AudioBench~\cite{wang-etal-2025-audiobench} and UltraEval-Audio~\cite{shi-etal-2026-ultraeval} sit alongside benchmarks such as LibriSpeech~\cite{7178964}, CoVoST2~\cite{wang21s_interspeech}, and VoiceBench~\cite{chen-etal-2026-voicebench}.
Each of these tools covers part of the modality space; evaluating an omni-modal model still means combining several of them.

\paragraph{Answer verification vs.\ preference judging.}
\emph{LLM-as-a-judge}~\citep{li2024llmsasjudgescomprehensivesurveyllmbased} asks a model to rate the quality of a response, often without a reference answer.
A \emph{verifier} solves a narrower problem: given the question, the reference answer, and the prediction, decide whether the prediction is correct.
Existing verifiers were built to check the answers of reasoning models on text benchmarks, or to provide verifiable rewards for reinforcement learning~\cite{chen2025xverifyefficientanswerverifier,liu-etal-2025-compassverifier,zhang2025generative}.
OmniEvaluator instead uses a verifier as part of its evaluation infrastructure, applying one lightweight, text-based model to benchmarks from all four modalities (\S\ref{sec:verifier}).

\paragraph{Gap addressed by this work.}
Many studies show how fragile evaluation scores can be: prompt changes~\cite{mizrahi-etal-2024-state,sclar2024quantifying}, the choice of evaluation examples~\cite{pmlr-v235-maia-polo24a}, answer-choice ordering~\cite{pezeshkpour-hruschka-2024-large}, and cross-framework differences~\cite{zhang-etal-2025-lmms,alzahrani-etal-2024-benchmarks} can each shift results substantially.
OmniEvaluator responds at the system level: a shared intermediate schema makes heterogeneous evaluators interoperable, every run yields a reproducible artifact, the verifier score holds steady when engine or prompt configurations differ, and a federated pipeline shares GPU servers across evaluations.
To our knowledge, no existing framework offers this combination.

\section{OmniEvaluator}

OmniEvaluator is a unified evaluation framework that centers all factors influencing evaluation outcomes---prompt template, generation configuration, model revision, benchmark version, and metric parameters---within a single, inspectable configuration.
The framework is guided by two design goals.
\emph{Multi-modal generality}: support benchmarks across text, image, video, audio, and cross-modality settings in a single pipeline, primarily by integrating entire evaluation frameworks---rather than individual benchmarks---to leverage existing implementations and keep pace with upstream releases; benchmarks that external frameworks do not cover, such as audio, omni-modal, and tool-calling tasks, are supported through a built-in engine.
\emph{Visualization and interpretability}: enable consistent cross-modal model comparison despite heterogeneous metric systems, with an integrated dashboard that visualizes capability profiles and trade-offs across modalities.

\begin{figure*}[t!]
  \centering
  \includegraphics[width=0.48\textwidth]{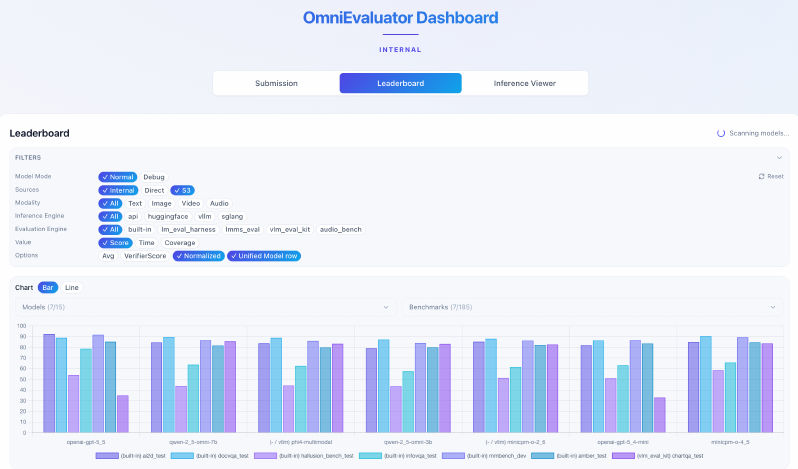}\hfill
  \includegraphics[width=0.48\textwidth]{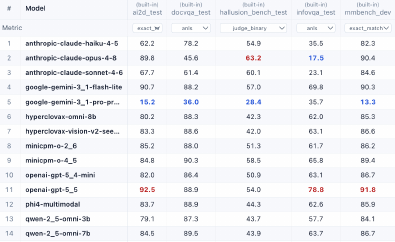}
  \caption{\textbf{OmniEvaluator dashboard (leaderboard view).}
  The dashboard synchronizes with evaluation artifacts produced by OmniEvaluator, providing an integrated view of benchmark results across experiments.
  \emph{Left}: composable filters (modality, inference engine, evaluation engine) and cross-model benchmark comparison charts.
  \emph{Right}: the consolidated leaderboard; missing entries explicitly surface modality coverage gaps (\url{https://github.com/naver-ai/omni-evaluator}).
  }
  \label{fig:dashboard}
\end{figure*}

\subsection{Composable Architecture through Intermediate Schema}
\label{sec:architecture}

OmniEvaluator decomposes evaluation into four independent stages---data iteration, inference, postprocessing, and metric computation---and mediates all inter-stage exchange through a unified intermediate schema (Figure~\ref{fig:architecture}).
Each record in this schema is a structured object containing the benchmark sample, the raw model prediction, the postprocessed output, and the computed metric score.
Below is one such record, for a video benchmark.
\begin{lstlisting}[basicstyle=\ttfamily\scriptsize]
{
  "benchmark": "mvbench_test_64frames",
  "messages": [{
    "role": "user",
    "content": [
      { "type": "video",
        "value": ".../action_sequence__0.mp4",
        "num_frames": 64,
        "sampling_strategy": "uniform" },
      { "type": "text",
        "value": "What happened after ...?" }
    ]
  }],
  "label": ["A"],
  "output": {
    "text": {
      "prediction": ["A. Ate the medicine."],
      "prediction_postprocessed": ["A"]
    },
    "reasoning_content": null
  },
  "metrics": { "exact_match": 1.0 }
}
\end{lstlisting}
The record shape does not change with modality: a text, image, video, or audio sample differs only in the \texttt{content} entries it carries and the modality-specific fields those entries add---\texttt{num\_frames} and \texttt{sampling\_strategy} for video, a sampling rate for audio---which the inference engine reads and the evaluation framework never has to know about.
Postprocessing writes to its own slot beside the raw prediction, so a metric can read either one without knowing which transformation produced it; the schema is described further in Appendix~\ref{app:schema}.
This common representation lets any inference engine feed into any postprocessor and metric module: the same multiple-choice extraction or ASR normalization logic is reused across all benchmarks of that task type, and shared metrics such as exact match are computed by a single implementation regardless of which framework defined the benchmark.

Each inference engine and evaluation framework listed in Table~\ref{tab:supported_engines} is connected to this schema through a thin adapter that translates between its native data format and the common representation, reducing integration cost from $O(N \!\times\! M)$ to $O(N \!+\! M)$: a single adapter makes any newly released upstream framework composable with all existing engines and modules.
Because postprocessing and metric modules are shared across frameworks, scoring discrepancies that would otherwise arise from differing \emph{metric implementations}---such as two frameworks reporting ``ANLS'' with different edit-distance thresholds---are structurally eliminated.
Full per-modality evaluation results obtained by combining these components are provided in Table~\ref{tab:leaderboard} (Appendix).

\subsection{Evaluation as a Reproducible Specification}
\label{sec:spec}

Every evaluation run emits a self-contained artifact that bundles: (i)~the evaluation configuration (prompt template, generation parameters, model revision, benchmark version, and metric settings), (ii)~all intermediate outputs (raw predictions and postprocessed results), and (iii)~the final scores.
This artifact is the unit of reproducibility: given the same artifact, any user can re-run the identical evaluation or inspect every decision that influenced a score; in practice, teams use artifacts as baseline anchors across training stages, checkpoint references for regression testing, and verifiable records for external releases.
The dashboard (\S\ref{sec:dashboard}) automatically ingests these artifacts, enabling cross-experiment comparison without manual data wrangling.

\subsection{Quick Start}
\label{sec:quickstart}

OmniEvaluator supports local CLI evaluation for fine-grained control and a remote mode in which users submit requests to a persistent server.
Both modes produce the same provenance-rich artifacts (\S\ref{sec:spec}) and stream results to the dashboard on completion; installation and server-launch details are given in Appendix~\ref{app:quickstart}.

\paragraph{Local evaluation.}
Users run evaluation directly via the CLI; an existing artifact can also be given as input to reproduce its results exactly:

\begin{lstlisting}[basicstyle=\ttfamily\scriptsize]
python run.py \
  --inference_engine=vllm \
  --url={vllm_url} \
  --exp_name=hyperclovax-seed-4b \
  --evaluation_engine=builtin \
  --benchmarks=haerae_vision_test
\end{lstlisting}

\paragraph{Remote evaluation.}
Users submit an evaluation request to a persistent server via a single REST call, with no local installation required---useful when a model undergoes multi-stage training and evaluation environments change frequently:

\begin{lstlisting}[basicstyle=\ttfamily\scriptsize]
curl -X POST http://{host}:{port}/add_job \
 -H "Content-Type: application/json" \
 -d '{"arguments": {
  "inference_engine": "huggingface",
  "model_name_or_path": "Qwen/Qwen2.5-Omni-3B",
  "exp_name": "qwen2.5-omni-3b",
  "evaluation_engine": "builtin",
  "benchmarks": "librispeech_test_clean"
 }}'
\end{lstlisting}

The server records each request with its evaluation specification (\S\ref{sec:spec}) for server-side reproducibility; results appear on the dashboard upon completion.

\begin{table}[ht]
  \centering
  \caption{\textbf{Verification accuracy on our human-verified held-out test split ($n{=}1{,}566$).} All models use an identical text-only configuration. Per-group best in \textbf{bold}, second \underline{underlined}.}
  \label{tab:verifier_main}
  \small
  \setlength{\tabcolsep}{6pt}
  \begin{adjustbox}{max width=\columnwidth}
  \begin{tabular}{llc}
    \toprule
    \textbf{Type} & \textbf{Verifier} & \textbf{acc} \\
    \midrule
    \multirow{6}{*}{\makecell[l]{Proprietary\\(API)}}
      & GPT-5.5                & \underline{90.2} \\
      & GPT-5.4-mini           & 82.8 \\
      & Claude-Opus-4.8        & \textbf{90.4} \\
      & Claude-Haiku-4.5       & 84.3 \\
      & Gemini-3.1-Pro         & 89.8 \\
      & Gemini-3.1-Flash-Lite  & 83.5 \\
    \midrule
    \multirow{6}{*}{\makecell[l]{Open-\\source}}
      & OmniEval Verifier      & \textbf{85.0} \\
      & Qwen3.5-9B             & \underline{79.9} \\
      & Qwen3.5-4B             & 79.8 \\
      & Qwen3.5-0.8B           & 64.2 \\
      & Qwen3-4B-Instruct-2507 & 76.2 \\
      & Qwen3-0.6B             & 56.1 \\
    \bottomrule
  \end{tabular}
  \end{adjustbox}
\end{table}

\begin{table*}[t]
  \centering
  \caption{
  \textbf{API-equivalent cost of one full evaluation pass.} Avg.\ In/Out Tok.\ are the mean input/output tokens per sample. Cost is in USD under public pricing.
  }
  \label{tab:verification_cost_api}
  \scriptsize
  \setlength{\tabcolsep}{4pt}
  \begin{adjustbox}{max width=\textwidth}
  \begin{tabular}{l rr rr rrrrrr}
    \toprule
    \textbf{Modality} & \textbf{\# Bench.} & \textbf{\# Samples} & \textbf{Avg.\ In Tok.} & \textbf{Avg.\ Out Tok.} & \textbf{GPT-5.4-mini} & \textbf{GPT-5.5} & \textbf{Claude-Haiku-4.5} & \textbf{Claude-Opus-4.8} & \textbf{Gemini-3.1-Flash-Lite} & \textbf{Gemini-3.1-Pro} \\
    \midrule
    Image & 8 & 21{,}433  & 44.3  & 29.2  & 3.53  & 23.52  & 4.08 & 20.39  & 7.06 & 16.94 \\
    Audio & 8 & 33{,}293  & 23.1  & 11.1  & 2.24  & 14.93 & 2.62  & 13.08  & 4.48 & 10.75 \\
    Video & 8 & 31{,}260  & 69.1  & 15.2  & 3.76  & 25.06  & 4.54  & 22.68  & 7.52 & 18.04 \\
    Text  & 8 & 43{,}640  & 133.8 & 177.0 & 39.14 & 260.92 & 44.46 & 222.30 & 78.28 & 187.86 \\
    \midrule
    \textbf{Total} & \textbf{32} & \textbf{129{,}626} & \textbf{75.0} & \textbf{70.9} & \textbf{48.67} & \textbf{324.43} & \textbf{55.69} & \textbf{278.60} & \textbf{97.34} & \textbf{233.59} \\
    \bottomrule
  \end{tabular}
  \end{adjustbox}
\end{table*}

\section{Supported Features}

\subsection{Integrated Dashboard}
\label{sec:dashboard}

The dashboard provides an integrated view of benchmark results across experiments and modalities (Figure~\ref{fig:dashboard}); evaluation artifacts are automatically synchronized via a remote storage backend (\S\ref{sec:spec}), so results become browsable as soon as a run completes.
Rather than crowning a single best model, the dashboard is designed to visualize cross-modal capability profiles and trade-offs, and it explicitly surfaces \emph{modality coverage gaps}: benchmarks a model has not been evaluated on appear as missing entries, making it immediately apparent which modalities remain untested.

The dashboard also offers interactive visualizations---including radar charts that overlay multiple models on user-selected benchmark axes---and side-by-side inspection of inference samples for contrasting predictions across training stages or against baselines to identify modality-specific error patterns (demonstrated in the live demo).
To support model selection, it further reports the mean of the \textbf{verifier score}---a single normalized value that our built-in verifier assigns to every benchmark on the same $[0,100]$ scale (\S\ref{sec:verifier})---across benchmarks: because it varies far less across evaluation engines and prompt conditions, its average is comparable where raw native metrics are not (measured on disparate, unbounded scales), and visualizing this single signal across models and checkpoints turns per-benchmark results into an actionable decision aid.

\subsection{Verifier}
\label{sec:verifier}

Rule-based metrics such as exact match, WER~\citep{park-etal-2024-automatic}, and BLEU~\citep{papineni-etal-2002-bleu} compare strings, and two situations complicate this comparison.
First, the score depends on how the evaluation is run: the same model outputs, scored under different prompt and metric configurations, swing by up to $88$ points on a $[0,100]$ scale (Table~\ref{tab:engine_comparison}).
Each framework's benchmark-specific prompt constrains the output shape---``a single word or phrase'', ``the option's letter''---and its metric implementation is written for that shape (Table~\ref{tab:prompt_conditions}), so the prompt is part of the harness rather than a neutral wrapper: when a model does not follow it closely, the metric no longer measures what it was written to measure.
How much this matters depends on the parser, which also differs across engines: on OCRBench the two frameworks disagree by more than $50$ points under the same prompt, while on RealWorldQA both parsers accept the answer in either form and the prompt barely matters.
Second, even when a model follows the prompt exactly, string comparison misses answers that are semantically correct but phrased differently from the reference or buried in reasoning traces, and a corpus-level average can be dominated by a few such cases.
A common remedy is to have an external LLM API judge the responses, but this exposes internal data to a third party, drifts as the judge version changes, and accumulates cost over repeated evaluations (Table~\ref{tab:verification_cost_api}).

Our remedy is to build the semantic check into the system itself.
We train and release \textbf{OmniEval Verifier}, a compact model that reads a \texttt{(question, reference, prediction)} text triple and returns a rationale and a binary verdict on whether the prediction answers the question correctly.
Training and data details are in Appendix~\ref{app:verifier_recipe}.
Built on \texttt{Qwen3-0.6B} and distributed as an 8-bit (Q8) GGUF model, it runs on CPU-only machines via \texttt{llama.cpp}\footnote{\url{https://github.com/ggml-org/llama.cpp}}, so OmniEvaluator ships it as a default scorer alongside native metrics, adding no external API calls and near-zero marginal cost.
Per-sample verdicts aggregate into the \textbf{verifier score}, a value on a $[0,100]$ scale computed the same way for every benchmark; on the outputs of Table~\ref{tab:engine_comparison}, this score moves far less than the native metric---on GQA and POPE, by a few points where the native metric moves by $40$ to $88$.
Under a benchmark-specific prompt the two largely agree (the \emph{Sp.}\ columns); the verifier's value is therefore not that it outperforms a well-configured native metric, but that it holds steady when the configuration does not match (the \emph{Un.}\ columns).

On our human-verified held-out test split, OmniEval Verifier reaches $85.0$ accuracy, matching or exceeding cost-efficient proprietary judges (GPT-5.4-mini $82.8$, Claude-Haiku-4.5 $84.3$) and outperforming the open-source models we evaluated under the same text-only configuration (Table~\ref{tab:verifier_main}).
Against native metrics, the verifier score closely tracks exact-match-style metrics, and for unbounded metrics such as WER and BLEU it adds a per-sample correct/incorrect view that corpus-level averages do not; where the two disagree, the gap is informative rather than contradictory, since a corpus-level average and a per-utterance verdict measure different things (Table~\ref{tab:native_vs_verifier}).
We position the verifier score as a \emph{complement} to native metrics rather than a replacement: native metrics remain the reference under their intended setup, while the verifier adds a signal that survives cross-engine and cross-prompt variation.
During model development, its mean proved comparable across benchmarks and training stages and aided model-selection decisions.

\begin{table}[t]
\centering
\caption{\textbf{Comparison of native benchmark metrics and Verifier Score.}
Native metrics are EM (exact match) for MMLU-Pro and GSM8K-CoT, PASS@1 for MBPP, WER (\%, $\downarrow$) for LibriSpeech, BLEU-4 ($\uparrow$) for CoVoST2 (en$\to$zh), and EM for VocalSound; the Verifier Score is uniformly scaled to $[0,100]$ ($\uparrow$).
Across these heterogeneous native scales, the Verifier Score provides a single comparable signal.}
\label{tab:native_vs_verifier}
\resizebox{\columnwidth}{!}{%
\begin{tabular}{lll l cc}
\toprule
\textbf{Modality} & \textbf{Benchmark} & \textbf{Model} & \textbf{Native Metric} & \textbf{Native} & \textbf{Verifier Score} \\
\midrule
\multirow{9}{*}{Text}
  & \multirow{3}{*}{MMLU-Pro}  & Qwen2.5-Omni-3B     & \multirow{3}{*}{EM ($\uparrow$)} & 42.0 & 41.9 \\
  & &                             Qwen2.5-Omni-7B     &                              & 51.1 & 51.0 \\
  & &                             HyperCLOVAX-SEED-4B &                              & 56.8 & 57.1 \\
\cmidrule(lr){2-6}
  & \multirow{3}{*}{MBPP}      & Qwen2.5-Omni-3B     & \multirow{3}{*}{PASS@1 ($\uparrow$)} & 53.4 & 83.6 \\
  & &                             Qwen2.5-Omni-7B     &                              & 50.4 & 86.5 \\
  & &                             HyperCLOVAX-SEED-4B &                              & 54.2 & 76.2 \\
\cmidrule(lr){2-6}
  & \multirow{3}{*}{GSM8K-CoT} & Qwen2.5-Omni-3B     & \multirow{3}{*}{EM ($\uparrow$)} & 32.4 & 81.3 \\
  & &                             Qwen2.5-Omni-7B     &                              & 70.2 & 84.8 \\
  & &                             HyperCLOVAX-SEED-4B &                              & 48.3 & 51.2 \\
\midrule
\multirow{9}{*}{Audio}
  & \multirow{3}{*}{LibriSpeech}       & Qwen2.5-Omni-3B  & \multirow{3}{*}{WER ($\downarrow$)}  & 2.6  & 71.2 \\
  & &                                    Qwen2.5-Omni-7B  &                                     & 2.3  & 63.5 \\
  & &                                    Phi-4-Multimodal &                                     & 1.7  & 82.9 \\
\cmidrule(lr){2-6}
  & \multirow{3}{*}{CoVoST2 (en$\to$zh)} & Qwen2.5-Omni-3B  & \multirow{3}{*}{BLEU-4 ($\uparrow$)}   & 0.0  & 35.4 \\
  & &                                    Qwen2.5-Omni-7B  &                                     & 0.0  & 37.1 \\
  & &                                    Phi-4-Multimodal &                                     & 0.0  & 36.1 \\
\cmidrule(lr){2-6}
  & \multirow{3}{*}{VocalSound}        & Qwen2.5-Omni-3B  & \multirow{3}{*}{EM ($\uparrow$)}        & 90.2 & 90.2 \\
  & &                                    Qwen2.5-Omni-7B  &                                     & 91.9 & 91.9 \\
  & &                                    Phi-4-Multimodal &                                     & 32.2 & 36.0 \\
\bottomrule
\end{tabular}%
}
\end{table}

\begin{table}[ht]
  \centering
  \caption{\textbf{Wall-time speedup of federated over conventional evaluation.} \textbf{Bold} / \underline{underline}: best / second-best per modality.}
  \label{tab:federated_evaluation_efficiency}
  \small
  \setlength{\tabcolsep}{5pt}
  \begin{adjustbox}{max width=\columnwidth}
  \begin{tabular}{lcccc}
    \toprule
    \textbf{Model} & Text & Image & Video & Avg. \\
    \midrule
    HyperCLOVAX-SEED-4B & 1.41$\times$ & 2.32$\times$ & \textbf{1.62}$\times$ & 1.78$\times$ \\
    HyperCLOVAX-SEED-Omni-8B & 1.28$\times$ & 2.29$\times$ & -- & 1.79$\times$ \\
    Qwen2.5-Omni-3B & \textbf{1.76}$\times$ & \underline{2.70}$\times$ & \textbf{1.62}$\times$ & \textbf{2.03}$\times$ \\
    Qwen2.5-Omni-7B & \underline{1.60}$\times$ & \textbf{2.80}$\times$ & \underline{1.54}$\times$ & \underline{1.98}$\times$ \\
    \midrule
    \textit{Modality avg.} & \textit{1.51}$\times$ & \textit{2.53}$\times$ & \textit{1.59}$\times$ & \\
    \bottomrule
  \end{tabular}
  \end{adjustbox}
\end{table}

\subsection{Federated Evaluation}
\label{sec:federated_evaluation}

The modular architecture of \S\ref{sec:architecture} separates inference from evaluation: models are hosted by serving engines such as vLLM, and evaluation clients talk to them over HTTP.
\textbf{Federated evaluation} builds on this split: inference servers and evaluation clients form a many-to-many pool, decoupling where benchmarks run from where GPUs are.
This decoupling yields two complementary benefits.
\emph{First, it improves utilization of fragmented GPU resources.}
A single client can spread benchmarks across multiple servers---a spare A100 on one machine, idle V100s on another---while data loading, metric computation, and verifier calls stay on CPU machines.
GPUs that would otherwise be too scattered to serve a single job are thus aggregated into one logical evaluation pool.
\emph{Second, it minimizes GPU idle time and thereby improves throughput.}
Many clients can drive one server, and their concurrent requests are merged by the engine's continuous (in-flight) batching, which keeps the accelerator saturated between requests rather than stalling on the request stream of any single evaluation process.
Under the same GPU allocation, these two effects together yield a $1.3$--$2.8\times$ wall-time speedup over conventional per-process evaluation, in which each framework runs inference from within its own process, with the largest gains on image benchmarks (Table~\ref{tab:federated_evaluation_efficiency}).

\section{Conclusion}
OmniEvaluator turns fragmented omni-modal evaluation into a single workflow: existing engines and frameworks compose through one schema, every run yields a reproducible artifact, and results flow into one dashboard.
Cross-framework comparisons (Table~\ref{tab:engine_comparison}) show why this matters: identically named benchmarks diverge across engines and prompt configurations, while the verifier score on the same predictions moves far less.
OmniEvaluator and the verifier are publicly released at \url{https://github.com/naver-ai/omni-evaluator} with a live demo and dashboard; a demo video is available at \url{https://www.youtube.com/watch?v=4Z5VZZWyXqY}.

\section*{Limitations}
OmniEvaluator wraps upstream evaluators rather than reimplementing them, so a bug in an upstream framework can flow into the scores it reports.
The design does not prevent this, but it makes such problems visible: every artifact pins exact framework versions, so re-running anchor models after an upgrade exposes score regressions (the daggered models in Table~\ref{tab:leaderboard}), and running the same benchmark under two engines surfaces disagreements that point to upstream issues (Table~\ref{tab:engine_comparison}).
The verifier judges only the textual triple and returns a binary verdict; these are the choices that keep it small enough for CPU and its score uniform across benchmarks, and its accuracy already matches or exceeds cost-efficient proprietary judges, though not the strongest ones (Table~\ref{tab:verifier_main}).
Consequently, it cannot assess criteria that go beyond textual correctness---such as visual grounding, audio quality, or generation quality---which some omni-modal tasks require; for these, users should fall back on the corresponding native metrics.
Finally, this paper covers omni-modal understanding, where predictions are text; extending evaluation and the verifier to multimodal outputs such as image and speech generation is a natural next step.


\section*{Acknowledgments}
We thank the anonymous reviewers and the program committee for their constructive feedback.
We are grateful to the Hyperscale AI team at NAVER Cloud AI for their work throughout the development of HyperCLOVA X 8B Omni and subsequent models, and for their support in building the evaluation setup this system grew out of.
We also thank Huiyeon Yang for help with the figures.

\bibliography{custom}

\clearpage
\appendix
\setcounter{dbltopnumber}{3}
\renewcommand{\dbltopfraction}{0.95}
\renewcommand{\topfraction}{0.95}
\renewcommand{\bottomfraction}{0.6}
\renewcommand{\textfraction}{0.03}
\renewcommand{\floatpagefraction}{0.95}

\section{Intermediate Schema}
\label{app:schema}

Every benchmark, regardless of modality, is represented by the same \texttt{Record} object (\S\ref{sec:architecture}).
Its \texttt{output} field holds a per-modality sub-output---\texttt{output.text} for text-producing benchmarks---carrying both the raw \texttt{prediction} and the \texttt{prediction\_postprocessed} written by the postprocess step, alongside a top-level \texttt{reasoning\_content} for models that emit a separate thinking trace.
Which fields carry data varies by benchmark, but the shape does not, and every engine and framework adapter reads and writes through the same slots.
Complete records for text, image, video, and audio benchmarks are available in the repository.\footnote{\url{https://github.com/naver-ai/omni-evaluator/tree/main/demo}}

\section{Installation and Server Launch}
\label{app:quickstart}

OmniEvaluator installs with a single command that clones the framework and its submodules; an administrator then starts the persistent evaluation server:

\begin{lstlisting}[basicstyle=\ttfamily\scriptsize]
git clone --recursive {repo}
cd OmniEvaluator && pip install -e .

python launch_server.py \
  --port {port} \
  --log_dir="./logs"
\end{lstlisting}



\section{Verifier: Data and Training}
\label{app:verifier_recipe}

\paragraph{Data.}
Training predictions are drawn from our own evaluation artifacts---the outputs of 16 evaluated models across four modalities and 155 benchmarks---to broaden the error distribution the verifier must score.
Gold labels come from a multi-teacher API pipeline to reduce single-teacher bias, balanced to a 1:1 positive/negative ratio with limited rule-based augmentation.
The held-out test split ($n{=}1{,}566$) is balanced over (modality, task) cells, prioritizes dispute samples, and is fully human-verified and disjoint from training.

\paragraph{Training.}
The verifier takes a \texttt{(question, reference, prediction)} triple and produces a rationale and a binary verdict (\texttt{Explanation: …\textbackslash nRating: 0|1}); the training mixture spans four modalities, from short exact-match responses to long reasoning traces.
Full training hyperparameters are in Table~\ref{tab:verifier_train_config}.

\paragraph{Failure modes of rule-based scoring.}
Corpus-level metrics can be dominated by a few pathological samples: on LibriSpeech, a refusal answer yields per-sample WER 266.6 and an off-transcript hallucination 175.0, distorting the aggregate, whereas the verifier simply marks such predictions incorrect.

\paragraph{Cost of API-based verification.}
Repeated over checkpoints and configurations, API-based verification compounds: at the per-pass cost in Table~\ref{tab:verification_cost_api}, $50{\times}5$ passes come to ${\approx}\$12{,}000$ for GPT-5.4-mini and ${\approx}\$70{,}000$ for Claude-Opus-4.8, whereas OmniEval Verifier runs at near-zero marginal cost.

\paragraph{Verification accuracy.}
On the human-verified test split, zero-shot inference with the base \texttt{Qwen3-0.6B} reaches only $56.1$ accuracy; trained as OmniEval Verifier---small enough to run on CPU alone---it rises to $85.0$, matching or exceeding cost-efficient proprietary judges (GPT-5.4-mini $82.8$, Claude-Haiku-4.5 $84.3$) and every open-source model we evaluated (Table~\ref{tab:verifier_main}).

\begin{table}[!bt]
  \centering
  \caption{OmniEval Verifier training hyperparameters.}
  \label{tab:verifier_train_config}
  \renewcommand{\arraystretch}{0.92}\scriptsize
  \begin{adjustbox}{max width=\columnwidth}
  \begin{tabular}{l p{0.52\columnwidth}}
    \toprule
    \textbf{Hyperparameter} & \textbf{Value} \\
    \midrule
    Base model            & Qwen3-0.6B \\
    LoRA $r$ / $\alpha$ / dropout & 8 / 16 / 0.05 \\
    LoRA targets          & LM decoder q,k,v,o,gate,up,down \\
    Trainable params      & $\approx$5.05M \\
    Supervision           & completion-only (target tokens) \\
    \midrule
    Epochs                & $\le$3 (best on validation) \\
    Learning rate         & 1e-4 \\
    Scheduler / warmup    & cosine / 0.1 \\
    Effective batch       & 2 $\times$ 8 (accum) $\times$ 8 GPU $= 128$ \\
    Gradient clip         & 1.0 \\
    Precision / quant.    & fp32 / none \\
    \texttt{max\_seq\_len} & 4096 \\
    \midrule
    Hardware              & 8$\times$V100 (32GB) \\
    Distributed           & DeepSpeed ZeRO-2 \\
    Seed                  & 42 \\
    \bottomrule
  \end{tabular}
  \end{adjustbox}
\end{table}

\begin{table*}[t]
  \centering
  \caption{\textbf{How the task prompt changes the answer format, and the answer format changes the score.} Per benchmark and engine: the instruction under benchmark-specific (\emph{Sp.}) and uniform (\emph{Un.}) prompts, one Qwen2.5-Omni-7B prediction, and its native/verifier score ($\{0,1\}$). Table~\ref{tab:engine_comparison} reports all five benchmarks.}
  \label{tab:prompt_conditions}
  \renewcommand{\arraystretch}{1.05}
  \scriptsize
  \setlength{\tabcolsep}{2pt}
  \begin{tabular}{@{}l l c p{4.0cm} p{1.8cm} p{4.6cm} c c@{}}
    \toprule
    \textbf{Bench.} & \textbf{Eval. Engine} & \textbf{Setting} & \textbf{Task prompt} & \textbf{Reference} & \textbf{Prediction} & \textbf{Native} & \textbf{Verifier} \\
    \midrule
    \multirow{4}{*}{GQA}
      & \multirow{2}{*}{lmms-eval}
        & Sp. & \texttt{Answer the question using a single word or phrase.}
              & \multirow{2}{*}{\texttt{aluminum}} & \texttt{aluminum} & 1 & 1 \\
      & & Un. & \texttt{Please Answer the question in an appropriate format.}
              & & \texttt{The fence is made of aluminum.} & 0 & 1 \\
      \cmidrule(l){2-8}
      & \multirow{2}{*}{VLMEvalKit}
        & Sp. & \texttt{Answer the question using a single word or phrase.}
              & \multirow{2}{*}{\texttt{brown}} & \texttt{brown} & 1 & 1 \\
      & & Un. & \texttt{Please Answer the question in an appropriate format.}
              & & \texttt{The ground in the picture is brown.} & 0 & 1 \\
    \midrule
    \multirow{4}{*}{POPE}
      & \multirow{2}{*}{lmms-eval}
        & Sp. & \texttt{Answer the question using a single word or phrase.}
              & \multirow{2}{*}{\texttt{yes}} & \texttt{Yes} & 1 & 1 \\
      & & Un. & \texttt{Please Answer the question in an appropriate format.}
              & & \texttt{Yes, there is a snowboard in the image. The person in the image is riding a snowboard down a snowy slope.} & 0 & 1 \\
      \cmidrule(l){2-8}
      & \multirow{2}{*}{VLMEvalKit}
        & Sp. & \emph{(inline)} \texttt{Please answer yes or no.}
              & \multirow{2}{*}{\texttt{Yes}} & \texttt{yes} & 1 & 1 \\
      & & Un. & \emph{(inline)} \texttt{Please Answer the question in an appropriate format.}
              & & \texttt{<points x1="112" y1="112" alt="bottle">bottle</points>} & 0 & 1 \\
    \bottomrule
  \end{tabular}
\end{table*}

\section{Supported Benchmarks and Full Results}
\label{app:leaderboard}

Table~\ref{tab:leaderboard} reports per-benchmark scores across text, image, video, and audio; full results are continuously updated on the OmniEvaluator dashboard.
Coverage spans general knowledge and instruction following, math, VQA and document understanding, video comprehension, speech recognition and translation, and sound and music understanding.

\begin{table*}[t!]
\centering
\setlength{\abovecaptionskip}{4pt}\setlength{\belowcaptionskip}{2pt}\renewcommand{\arraystretch}{0.95}
\caption{
\textbf{Per-modality benchmark results} (accuracy, \%; OCRBench 0--1{,}000; ASR/AST in WER $\downarrow$ / BLEU $\uparrow$). Representative models shown; full results on the OmniEvaluator dashboard. \textsuperscript{\dag}Anchor models, re-evaluated on every framework or config update to monitor regressions.
}
\label{tab:leaderboard}
\fontsize{7.5}{8.5}\selectfont
\setlength{\tabcolsep}{3pt}
\begin{tabular}{@{}l l cccc cccc@{}}
\toprule
\textbf{Mod.} & \textbf{Model} & \multicolumn{8}{c}{\textbf{Benchmarks}} \\
\midrule
\multirow{10}{*}{\rotatebox{90}{\fontsize{8}{9}\selectfont Text}}
 & & \rotatebox{90}{MMLU-Redux} & \rotatebox{90}{MMLU-Pro} & \rotatebox{90}{IFEval} & \rotatebox{90}{ARC-C} & \rotatebox{90}{MATH} & \rotatebox{90}{AIME'25} & \rotatebox{90}{HLE} & \rotatebox{90}{KoBALT} \\
\cmidrule(l){3-10}
 & GPT-5.5                  & 96.4 & 88.2 & 94.1 & 96.2 & 99.1 & 100.0 & 41.6 & 87.4 \\
 & GPT-5.4-mini             & 84.2 & 77.4 & 85.8 & 94.3 & 89.1 & 36.7  & 5.0  & 39.4 \\
 & Claude-Opus-4.8          & 93.3 & 88.9 & 85.0 & 96.2 & 98.3 & 90.0  & -    & -    \\
 & Gemini-3.1-Pro           & 96.8 & 74.4 & 50.1 & 96.5 & 98.8 & 40.0  & 20.3 & 87.1 \\
 & HyperCLOVAX-SEED-Omni-8B & -    & 52.3 & 69.1 & -    & 71.2 & -     & -    & -    \\
 & Qwen3-Omni-30B-Instruct\textsuperscript{\dag}   & 88.2 & -    & -    & 94.7 & 96.9 & 56.7  & 5.4  & 36.1 \\
 & Qwen2.5-Omni-7B\textsuperscript{\dag}  & 74.1 & 51.1 & 51.9 & 86.4 & 71.1 & 6.7   & 5.1  & 20.9 \\
 & Phi-4-Multimodal\textsuperscript{\dag} & 69.3 & 50.4 & 70.6 & 82.3 & 60.8 & 3.3   & 5.8  & 11.3 \\
 & MiniCPM-o-4.5            & 76.0 & 60.4 & 83.0 & 90.0 & 82.4 & 16.7  & 4.3  & -    \\
\midrule
\multirow{10}{*}{\rotatebox{90}{\fontsize{8}{9}\selectfont Image/Video}}
 & & \rotatebox{90}{MMBench} & \rotatebox{90}{MMMU} & \rotatebox{90}{MathVista} & \rotatebox{90}{AI2D} & \rotatebox{90}{DocVQA} & \rotatebox{90}{OCRBench} & \rotatebox{90}{V-MME} & \rotatebox{90}{MVBench} \\
\cmidrule(l){3-10}
 & GPT-5.5                  & 91.8 & 24.9 & 52.4 & 92.5 & 88.9 & 808 & 56.7 & 36.1 \\
 & GPT-5.4-mini             & 86.7 & 23.8 & 44.8 & 82.0 & 86.4 & 793 & 42.7 & 28.9 \\
 & Claude-Opus-4.8          & 90.4 & 55.9 & 50.8 & 89.8 & -    & 840 & 52.1 & 18.1 \\
 & Claude-Sonnet-4.6        & 84.6 & -    & 48.1 & 67.7 & 61.4 & 839 & 51.0 & 28.8 \\
 & HyperCLOVAX-SEED-Omni-8B & 85.3 & 38.8 & -    & 80.2 & 88.3 & 769 & -    & -    \\
 & Qwen3-Omni-30B-Instruct\textsuperscript{\dag}   & 89.9 & 42.2 & 51.8 & 87.6 & 91.6 & 774 & 69.2 & 36.1 \\
 & Qwen2.5-Omni-7B\textsuperscript{\dag}  & 86.7 & 47.2 & 59.8 & 84.5 & 89.5 & 847 & 61.1 & 68.2 \\
 & Phi-4-Multimodal\textsuperscript{\dag} & 85.9 & 47.7 & 61.2 & 83.7 & 88.9 & 821 & 33.6 & 35.5 \\
 & MiniCPM-o-4.5            & 89.4 & 48.2 & 51.8 & 84.8 & 90.3 & 832 & 71.3 & 63.3 \\
\midrule
\multirow{10}{*}{\rotatebox{90}{\fontsize{8}{9}\selectfont Audio}}
 & & \rotatebox{90}{\makecell{LibriSp.\\clean $\downarrow$}} & \rotatebox{90}{\makecell{LibriSp.\\other $\downarrow$}} & \rotatebox{90}{Fleurs $\downarrow$} & \rotatebox{90}{CV15 $\downarrow$} & \rotatebox{90}{\makecell{CoVoST2\\en-zh $\uparrow$}} & \rotatebox{90}{VocalSound} & \rotatebox{90}{ClothoAQA} & \rotatebox{90}{MuchoMusic} \\
\cmidrule(l){3-10}
 & Whisper-large-v3         & 3.96 & 2.01 & 4.03 & 24.35 & -   & -    & -    & -    \\
 & Qwen2-Audio-Instruct     & 38.2 & 38.2 & 52.8 & 29.4  & 0.1 & 78.6 & 62.4 & 18.6 \\
 & Voxtral-Small            & 32.7 & 35.2 & 35.9 & 22.0  & 5.1 & 45.7 & 53.1 & 43.4 \\
 & Voxtral-Mini             & 39.1 & 44.0 & 32.1 & 20.5  & 3.3 & 0.1  & 36.8 & 50.8 \\
 & HyperCLOVAX-SEED-Omni-8B & 2.6  & 4.5  & 7.6  & -     & -   & -    & -    & -    \\
 & Qwen3-Omni-30B-Instruct\textsuperscript{\dag}   & 2.8  & 1.7  & 3.4  & 11.7  & 2.8 & 90.8 & 75.0 & 78.2 \\
 & Qwen2.5-Omni-7B\textsuperscript{\dag}  & 4.3  & 4.3  & 5.0  & 15.4  & 2.5 & 85.4 & 73.6 & 75.5 \\
 & Phi-4-Multimodal\textsuperscript{\dag} & 28.5 & 30.7 & 29.3 & 13.1  & 7.2 & 27.8 & 48.9 & 48.1 \\
 & MiniCPM-o-4.5            & 3.9  & 1.7  & 3.9  & 13.3  & 0.7 & 74.8 & 54.1 & 73.4 \\
\bottomrule
\end{tabular}
\end{table*}

\end{document}